\documentclass[11pt]{article}

\usepackage{graphicx}     
\usepackage{algorithm}
\usepackage{algpseudocode}
\usepackage{hyperref}
\usepackage{amsmath}
\usepackage{lineno}

\begin{document}

\title{Generating readily synthesizable small molecule fluorophore scaffolds with reinforcement learning}

\author{
  Ruhi Sayana\thanks{Department of Computer Science, Stanford University. Equal Contribution.} \and
  Kate Callon \footnotemark[1] \and
  Jennifer Xu\footnotemark[1] \and
  Jonathan Deutsch\thanks{Department of Molecular and Cellular Physiology, Stanford University} \and
  Steven Chu\thanks{Department of Molecular and Cellular Physiology, Stanford University; Department of Physics, Stanford University.} \and
  James Zou\thanks{Department of Computer Science, Stanford University; 
  Department of Biomedical Data Science, Stanford University.} \and
  John Janetzko\thanks{Department of Molecular and Cellular Physiology, Stanford University; 
  Department of Biochemistry and Molecular Genetics, University of Colorado Anschutz Medical Campus; Department of Pharmacology, University of Colorado Anschutz Medical Campus; Department of Pharmaceutical Sciences, Skaggs School of Pharmacy, University of Colorado Anschutz Medical Campus; BioFrontiers Institute, University of Colorado Boulder. Corresponding author: john.janetzko@cuanschutz.edu} \and
  Rabindra V. Shivnaraine\thanks{Department of Molecular and Cellular Physiology, Stanford University; 
  Currently at Greenstone Biosciences. Corresponding author: rvshiv@stanford.edu.} \and
  Kyle Swanson\thanks{Department of Computer Science, Stanford University. Corresponding author: swansonk@stanford.edu.}
}

\date{}

\maketitle

\begin{abstract}
Developing new fluorophores for advanced imaging techniques requires exploring new chemical space. While generative AI approaches have shown promise in designing novel dye scaffolds, prior efforts often produced synthetically intractable candidates due to a lack of reaction constraints. Here, we developed SyntheFluor-RL, a generative AI model that employs known reaction libraries and molecular building blocks to create readily synthesizable fluorescent molecule scaffolds via reinforcement learning. To guide the generation of fluorophores, SyntheFluor-RL employs a scoring function built on multiple graph neural networks (GNNs) that predict key photophysical properties, including photoluminescence quantum yield, absorption, and emission wavelengths. These outputs are dynamically weighted and combined with a computed $\pi$-conjugation score to prioritize candidates with desirable optical characteristics and synthetic feasibility. SyntheFluor-RL generated 11,590 candidate molecules, which were filtered to 19 structures predicted to possess dye-like properties. Of the 19 molecules, 14 were synthesized and 13 were experimentally confirmed. The top three were characterized, with the lead compound featuring a benzothiadiazole chromophore and exhibiting strong fluorescence (PLQY = 0.62), a large Stokes shift (97 nm), and a long excited-state lifetime (11.5 ns). These results demonstrate the effectiveness of SyntheFluor-RL in the identification of synthetically accessible fluorophores for further development.
\end{abstract}

\section{Introduction}
Fluorescent dyes with highly optimized photophysical properties are critical for advanced imaging methods to detect and interrogate biomolecules \cite{suzuki2007recent, datta2021recent}. A variety of microscopy techniques, such as STORM (Stochastic Optical Reconstruction Microscopy), PALM (Photoactivated Localization Microscopy) \cite{lelek2021single, henriques2011palm}, single-molecule FRET (Fluorescence Resonance Energy Transfer)\cite{sasmal2016single}, single-particle tracking (SPT), light-sheet microscopy \cite{santi2011light}, multi-photon fluorescence microscopy\cite{gratton2001multiphoton}, and fluorescence recovery after photobleaching (FRAP), have significantly enhanced our ability to investigate cellular structures and processes with improved spatio-temporal resolution. As these methodologies evolve, the demand for a diverse array of fluorescent dyes—each characterized by unique photophysical properties tailored to each specific imaging technique—becomes increasingly critical. For example, techniques such as SPT would benefit from dyes that are longer lived, whereas FRAP would benefit from dyes that are rapidly bleached.

Four primary families of dye scaffolds dominate the field of bioimaging: coumarins, xanthenes (such as fluorescein and rhodamine), BODIPYs, and cyanines\cite{colas2021prevalent, samanta2023xanthene}. These scaffolds form the basis for 29 Alexa Fluor dyes, 50 ATTO dyes, 6 Cy dyes, and at least 15 Janelia Fluor dyes, collectively addressing a broad range of imaging needs. Additionally, alternative scaffolds such as pyrene, naphthalene, carbazole, oxazine, porphyrin, benzothia(dia)zole, squaraine, and phthalocyanine have found more limited applications \cite{makar2019naphthalene, s2016carbazole}. Advances in dye chemistry continue to drive the discovery of new scaffolds and the development of synthetic routes for novel dyes with tailored photophysical properties, enabling higher imaging resolution and compatibility with modern imaging modalities \cite{islam2024small}. 

Recently, machine learning models have emerged as transformative tools for designing novel dye scaffolds. In one approach, trained property predictors based on characteristics such as photoluminescence quantum yield (PLQY), absorption, and emission are used to identify promising fluorescent molecules within databases of known compounds 
\cite{wang2021development,ye2020predicting,ju2021chemfluor, huang2024ai, bu2023designing}. While this method effectively uncovers novel candidates, generative models offer the capability to rapidly explore a broader chemical space and create entirely new scaffolds that are absent from existing fluorophore libraries or databases. Despite the potential of generative AI approaches to discover novel fluorescent molecular structures \cite{sumita2022novo, tan2023novo, zhu2025modular}, they often face challenges due to the generation of synthetically intractable scaffolds.

In this study, we developed SyntheFluor-RL, a generative AI model, to assemble readily synthesizable fluorescent molecule scaffolds. SyntheFluor-RL, a variant of the generative model SyntheMol-RL \cite{synthemol-rl}, assembles novel molecules using building blocks and well-established synthetic reactions available in the Enamine REAL Space \cite{grygorenko2020generating}, which contains over 30 billion possible molecules. SyntheFluor-RL was specifically adapted for fluorescent molecule design by training graph neural network-based property predictor models on the ChemFluor dataset \cite{ju2021chemfluor} to predict photoluminescence quantum yield (PLQY) and absorption and emission wavelengths. These property predictors are used to guide the model in generating molecules that were optimized for multiple fluorescent properties, leveraging an additional 57 reaction pathways from the Enamine  REAL space relevant for fluorescent molecule design, along with the 13 Enamine REAL space reaction pathways used by SyntheMol-RL. SyntheFluor-RL generated 11,590 candidate molecules, which were subsequently filtered and screened for predicted properties, structural novelty, and chemical diversity to identify the top 19 compounds for synthesis and experimental validation. Among the 13 compounds that were successfully tested, our experiments identified three distinct molecules, each derived from a different chemical scaffold. This demonstrates the utility of SyntheFluor-RL for diverse fluorescent molecule scaffold design.

\section{Results}
\begin{figure}
  \includegraphics[width=\textwidth]{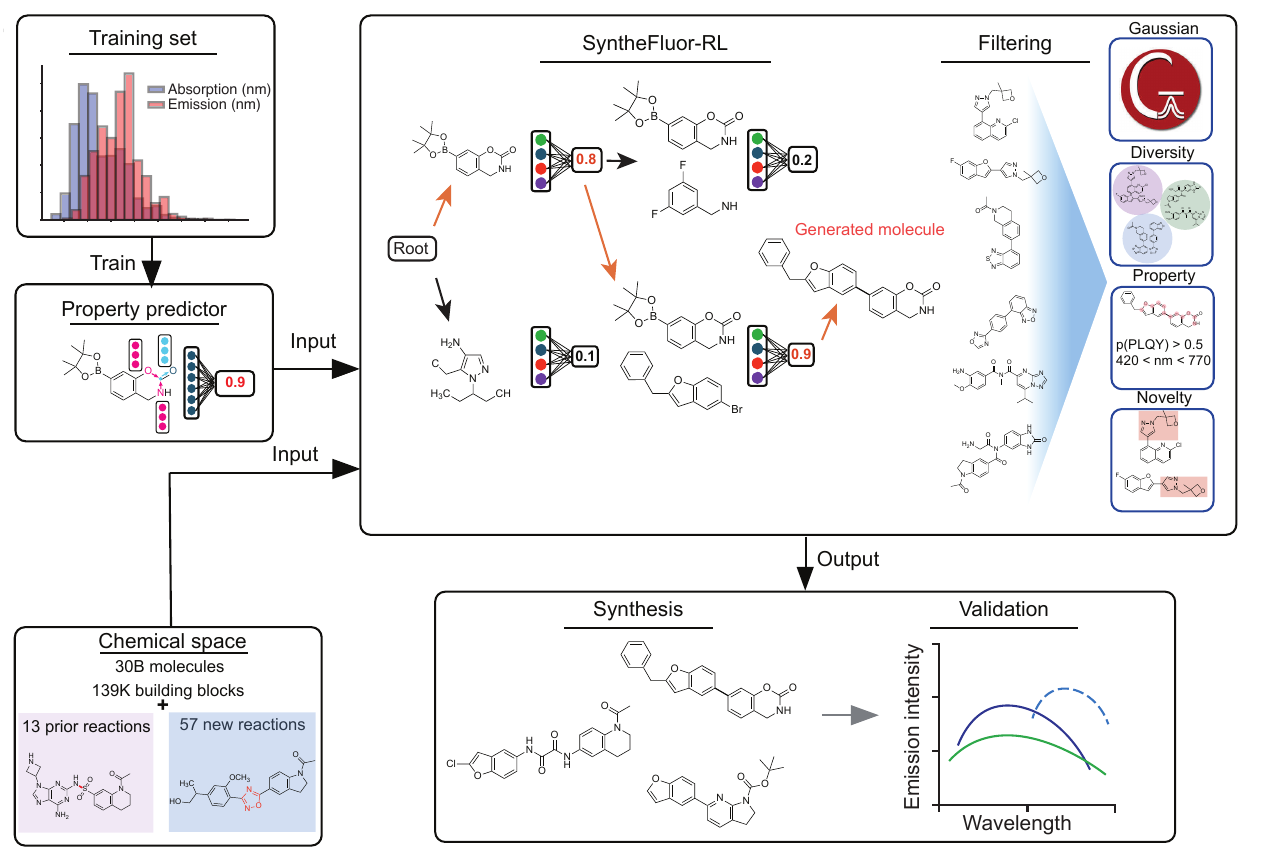}
  \centering
  \caption{\textbf{SyntheFluor-RL pipeline overview.} The workflow begins with training set processing followed by property predictor model development. Next, SyntheFluor-RL generates molecules using the trained property predictors within a chemical space of synthesizable molecules. These molecules are then filtered based on calculated and predicted properties, structural diversity, and novelty. Finally, selected molecules were synthesized and experimentally validated.
}
  \label{fgr:figure1}
\end{figure}

SyntheFluor-RL was applied to fluorescent molecule scaffold design via three main steps, as outlined in Figure \ref{fgr:figure1}: (1) training property predictor models on the ChemFluor dataset, (2) generating and filtering output molecules with SyntheFluor-RL, and (3) experimentally validating the output through synthesis and testing.

\subsection{Developing Property Prediction Models}
\begin{figure}
  \includegraphics[width=0.75\textwidth]{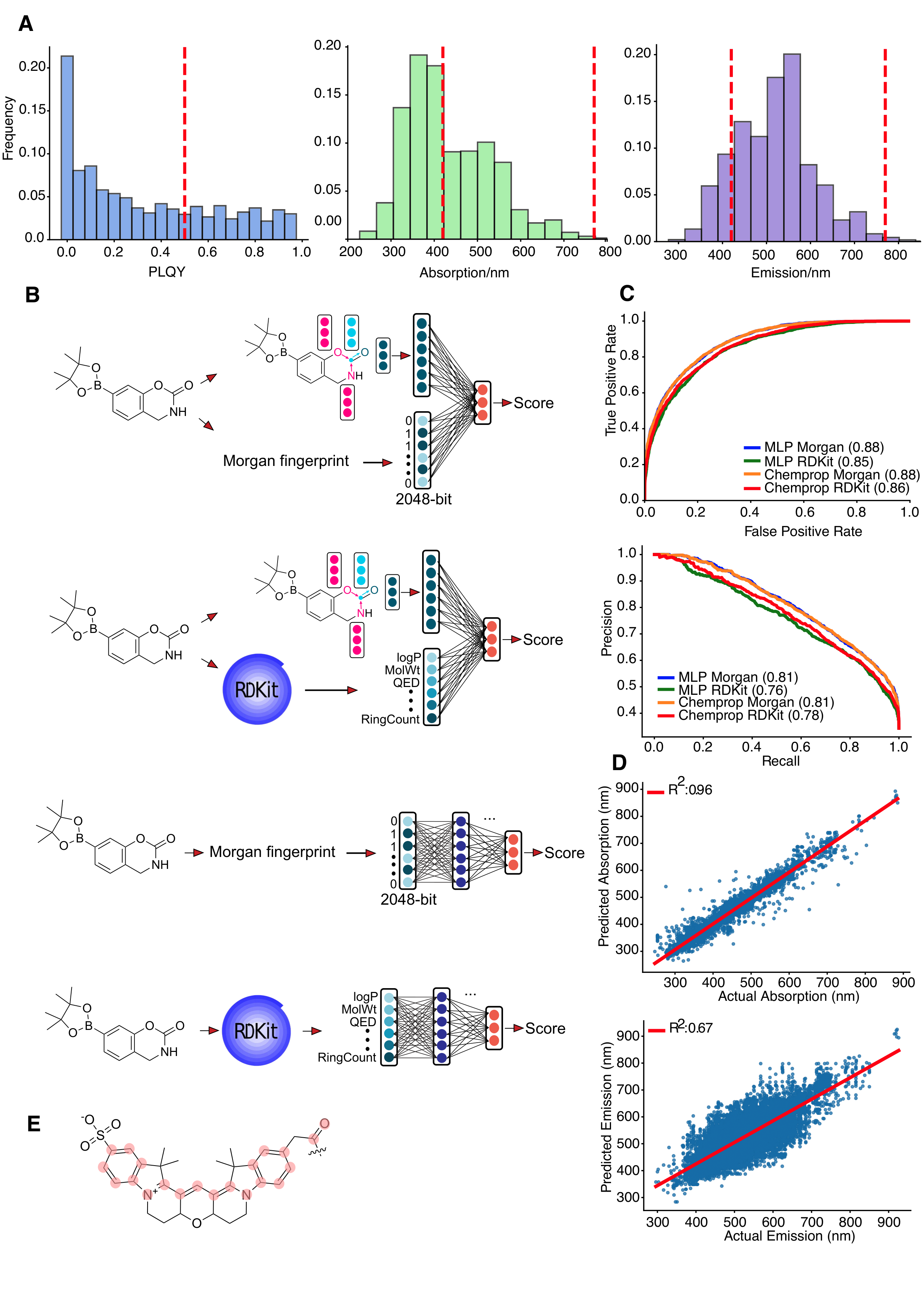}
  \centering
  \caption{\small \textbf{Property Prediction Model Development.}
  A) Distributions of photoluminescence quantum yield (PLQY), absorption wavelengths, and emission wavelengths in the ChemFluor dataset. Red lines represent cutoff values for PLQY classification model ($PLQY > 0.5$ considered fluorescent) and wavelengths considered to be in the visible spectra (420 nm to 770 nm) for absorption and emission.
  B) Visualizations of graph neural network architectures combined with either Morgan fingerprints or RDKit fingerprints (top two) and MLP architectures using Morgan or RDKit fingerprints (bottom two).
  C) ROC curves and PRC curves for PLQY classification models. 
  D) Plots of predicted absorption versus actual absorption for the absorption regression model using the Chemprop-Morgan architecture, and predicted emission vs actual emission for the emission regression model using the Chemprop-Morgan architecture.
  E) Visualization of sp$^2$ network algorithm. Atoms and bonds highlighted in red are members of the largest connected network of sp$^2$-hybridized atoms in the molecule.
}
  \label{fgr:figure2}
\end{figure}
The ChemFluor dataset \cite{ju2021chemfluor} was processed to extract PLQY, absorption, and emission values for 2,912 unique molecules dissolved in 63 different solvents, resulting in 4,336 unique molecule-solvent pairs (Figure \ref{fgr:figure2}A). The PLQY values ranged from 0 to 1, absorption values 247nm to 1026nm, and the emission values 296nm to 1045nm. Following curation, two machine learning model architectures were trained to predict fluorescence properties: (1) Chemprop (v1.6.1) \cite{yang2019analyzing}, a graph neural network that processes molecular graphs and computed features, and (2) a multilayer perceptron (MLP), which uses only computed molecular features (see Methods). Two types of features were tested for both models: Morgan fingerprints, which encode local chemical structures, and RDKit features, comprising 200 physicochemical properties computed with RDKit \cite{rdkit} (Figure \ref{fgr:figure2}B). In both cases, molecular features were augmented with four experimentally derived solvent properties—polarizability (SP), dipolarity (SdP), acidity (SA), and basicity (SB)—corresponding to the solvent used during experimental measurements.

Chemprop and MLP models were trained as either binary classifiers or regressors, depending on the task. PLQY prediction was modeled as a binary classification task using a threshold of $PLQY > 0.5$, while absorption and emission wavelength predictions were modeled as regression tasks. All models were trained using 10-fold cross-validation with an 80\% training, 10\% validation, and 10\% testing split, completing in under 60 minutes on an 8-CPU machine. PLQY models were trained on 3,055 molecule-solvent pairs, absorption models on 4,202 pairs, and emission models on 4,333 pairs.

Across all three tasks (PLQY, absorption, and emission) and both model architectures, the Morgan fingerprints outperformed the RDKit features (Figure \ref{fgr:figure2}C-D, Supplementary Table 1). The Chemprop-Morgan and MLP-Morgan architectures showed comparable performance on the PLQY classification task (Chemprop-Morgan: ROC-AUC = 0.895 $\pm$ 0.019 ; MLP-Morgan: ROC-AUC = 0.896 $\pm$ 0.019). Chemprop-Morgan demonstrated a slight advantage over MLP-Morgan for both absorption (Chemprop-Morgan MAE = 13.118 $\pm$ 1.203; MLP-Morgan MAE = 13.657 $\pm$ 1.083) and emission (Chemprop-Morgan MAE = 18.951 $\pm$ 0.986; MLP-Morgan MAE = 19.829 $\pm$ 1.268) regression tasks. Based on its better performance across most tasks, the Chemprop-Morgan architecture was selected for the scoring function in the molecule generation process.

The presence of a $\pi$-conjugated system is critical for a molecule to be fluorescent\cite{yamaguchi2008pi, zhang2023effect}, and larger $\pi$-conjugated systems reduce the HOMO-LUMO gap, shifting electronic transitions to the visible spectrum. To incorporate this information, we designed an algorithm (see Methods) to calculate the size of the largest network of connected atoms with sp$^2$ hybridization in each molecule’s molecular graph representation (Figure \ref{fgr:figure2}E).

\subsection{Generating Molecules with SyntheFluor-RL}
\begin{figure}
  \includegraphics[width=0.75\textwidth]{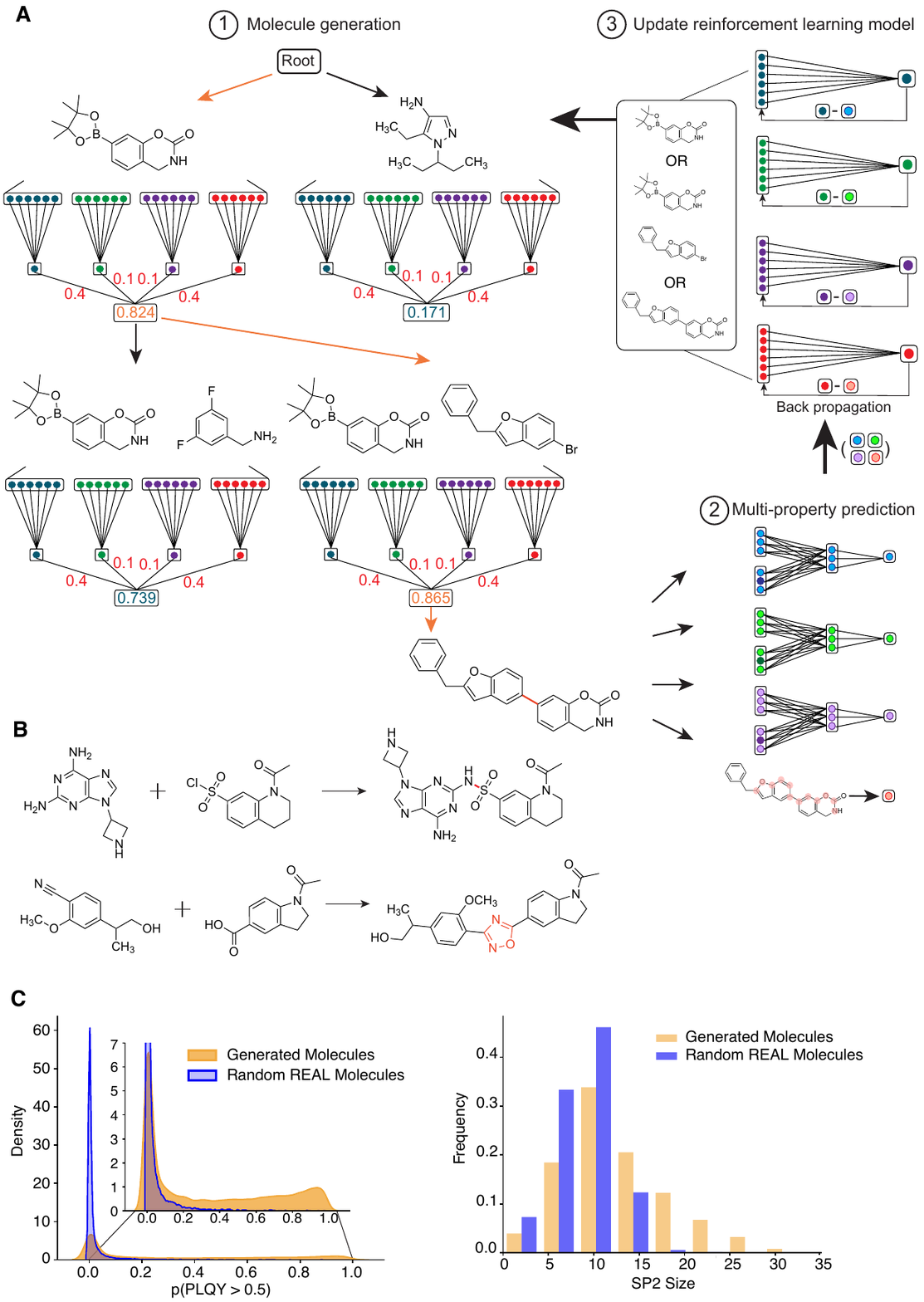}
  \centering
  \caption{\small \textbf{SyntheFluor-RL development and performance.} A) Schematic of the SyntheFluor-RL reinforcement learning algorithm. Step 1 shows the selection of building blocks with intermediate scoring conducted by four MLP-Morgan models (PLQY (blue), absorption (green), emission (purple), for sp$^2$ (red)), and the pairing of the final selected building blocks via Reaction 2718 to create the target molecule. Step 2 shows the evaluation of the candidate model via four models (Chemprop-Morgan for PLQY (light blue), absorption (light green), and emission (light purple), and the sp$^2$ algorithm for sp$^2$ network size (light red)). Step 3 shows how the scores from Step 2 are used to update the corresponding MLP models in Step 1 and re-weight the building blocks for the next rollout.
  B) An example of a non-ring forming reaction in the original set of 13 reactions (top), and a ring-forming reaction in the extended set of 70 reactions (bottom). 
  C) Distribution of PLQY probabilities (left) and histogram of sp$^2$ network sizes (right) on the generated molecules versus a random sample of 10,000 molecules in the REAL Space.
}
  \label{fgr:figure3}
\end{figure}
We developed SyntheFluor-RL (see Methods) to generate fluorescent molecule candidates (Figure \ref{fgr:figure3}A). SyntheFluor-RL is an adaptation of SyntheMol-RL \cite{synthemol-rl}, which is a generative model that uses reinforcement learning (RL) to design easy-to-synthesize molecule with desirable properties. SyntheMol-RL was previously applied to generate antibiotic candidates targeting \textit{Staphylococcus aureus}. Key to the SyntheMol-RL algorithm is its internal RL value function, which is used to direct its selection of chemical building blocks. Similarly, SyntheFluor-RL utilizes an RL value function to guide its selection process. However, its multi-parameter objective is altered to optimize for four essential properties: PLQY, absorption wavelength, emission wavelength, and sp$^2$ network size. 

During its generation process, SyntheFluor-RL computes a weighted score for each candidate molecule, with dynamically adjusted weights for each property to maximize the likelihood of generating molecules that satisfy all four criteria (Supplementary Figure 1A). Success criteria are defined for each of the four properties, and the weights are adjusted over time based on the rolling average success rates, ensuring a balanced optimization of photophysical properties. The trained Chemprop-Morgan property prediction models are used in the reward function that evaluates each candidate molecule, while an associated set of MLP-Morgan models are weighted in the value function to score intermediate candidates and are trained during generation. The MLP-Morgan architecture was chosen for the value function instead of a matching Chemprop-Morgan architecture because SyntheMol-RL with an MLP value function model is significantly faster than with a Chemprop value function. As a result, SyntheFluor-RL rapidly evaluates many combinations of building blocks using an MLP-Morgan model and then performs a slower but more accurate evaluation of generated molecules using a Chemprop-Morgan architecture. 

Notably, to ensure that fluorescence properties were assessed under relevant aqueous experimental conditions, all molecules evaluated by the Chemprop and MLP models within SyntheFluor-RL were represented as a concatenation of their Morgan fingerprint with four solvent features corresponding to water: polarizability (SP), dipolarity (SdP), acidity (SA), and basicity (SB).

Initial generations using SyntheFluor-RL yielded a scarcity of molecules with extended aromatic systems -- a critical feature for fluorescent dyes. This limitation arose from the lack of reactions that facilitate aromatic ring formation in the original set of 13 reactions in SyntheMol-RL. To address this, we added 57 new reactions from the Enamine REAL space to the reaction set---several of which specifically form aryl-aryl connections (e.g. Suzuki-Miyaura coupling) or construct new rings when joining two fragments ---bringing the total reaction count to 70 (Figure \ref{fgr:figure3}B).

Using this extended reaction set, SyntheFluor-RL executed 10,000 rollouts, completing the generation process in 16 hours, 38 minutes, and 26 seconds and generating 11,590 candidate fluorescent molecules. These generated molecules used 18 unique reactions, five of which were from the new set of reactions (Supplementary Figure 1B). Additionally, the PLQY property was weighted the most throughout the generation (Supplementary Figure 1A).

To evaluate SyntheFluor-RL's efficacy in generating molecules with optimized fluorescent properties, we compared the PLQY, absorption wavelengths, emission wavelengths, and sp$^2$ network sizes of the generated molecules against a random sample of 10,000 molecules from the Enamine REAL space. Molecules generated by SyntheFluor-RL had a higher probability of $PLQY > 0.5$ and larger sp$^2$ network sizes compared to the random sample (Figure \ref{fgr:figure3}C) while matching the overall absorption and emission wavelength distributions (Supplementary Figure 2), thereby demonstrating that SyntheFluor-RL successfully enriched for key fluorescent properties.

\begin{figure}
  \includegraphics[width=\textwidth]{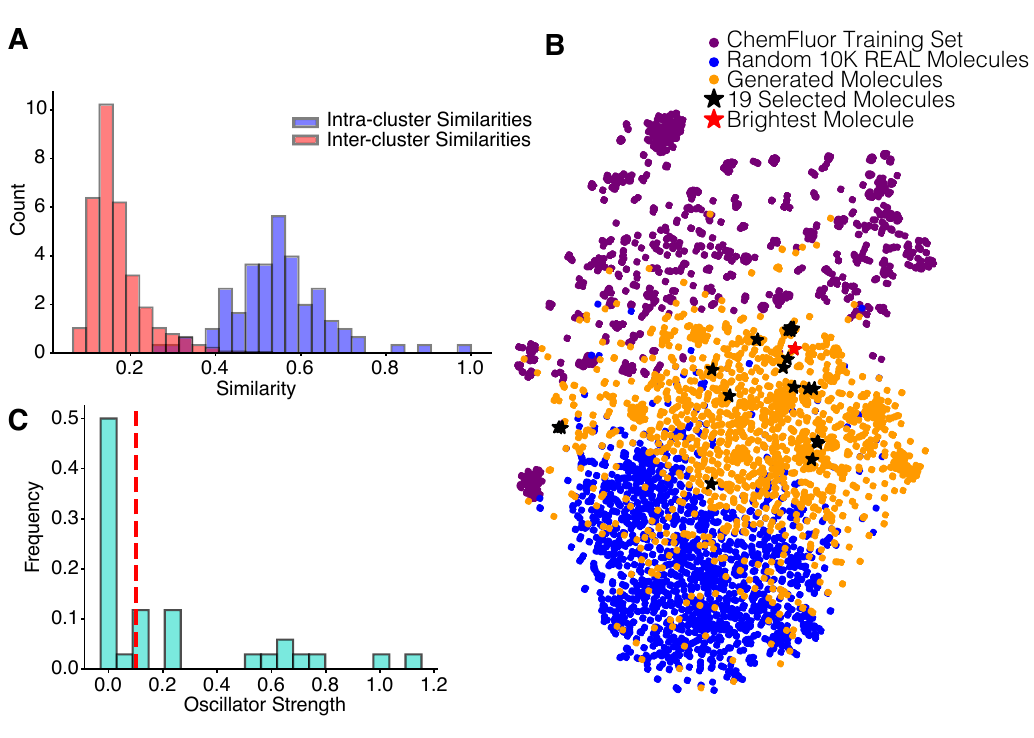}
  \centering
  \caption{\small \textbf{Molecule selection for synthesis.}
A) Histogram of intra-cluster and inter-cluster Tanimoto similarities for molecules separated into 100 clusters via K-means.
B) t-SNE representation of ChemFluor training set molecules, randomly selected molecules in REAL Space, generated molecules, and the final 19 selected molecules.
C) Histogram of calculated oscillator strength in the 52 selected molecules. Red line shows the 0.01 cutoff for oscillator strength.
}
  \label{fgr:figure4}
\end{figure}

To identify the most promising candidates, we applied a sequential multi-step filtering process. First, molecules with an sp$^2$ network size smaller than 12 were removed, excluding 5,479 molecules. Next, only molecules with a predicted probability of $PLQY > 0.5$ were retained, eliminating 4,256 molecules. Molecules with predicted absorption and emission wavelengths outside the visible range (420–750 nm) were also removed, excluding 21 molecules based on absorption and 1,203 based on emission. This left 631 molecules. To ensure structural diversity, we grouped these molecules into 100 clusters using K-means clustering on Morgan fingerprints using Tanimoto similarity (Figure \ref{fgr:figure4}A). Then, we manually selected one molecule per cluster to maintain diversity, yielding 52 candidates. Of these, 34 (65\%) were available for synthesis by Enamine.

Novelty of the generated molecules was calculated from the Tanimoto similarity between each generated molecule and each ChemFluor molecule. 630 of the 631 generated molecules that passed the PLQY, absorption, emission, and sp$^2$ filtering steps had Tanimoto similarities $< 0.5$ with ChemFluor molecules. A qualitative comparison was performed using t-SNE analysis on the Morgan fingerprints with Tanimoto similarity as the distance metric and PCA for initialization, comparing 2,000 samples each from Enamine REAL molecules, the ChemFluor dataset, and SyntheFluor-RL-generated molecules (Figure \ref{fgr:figure4}B). The t-SNE plot revealed that the generated molecules occupy a novel chemical space. Overlap between the generated molecules, the random real molecules, and the ChemFluor molecules also suggests that the generated molecules are realistic and may share fluorescent properties.

The filtering steps narrowed the potential fluorescent candidates from 11,590 molecules to 34 molecules. A final non-critical filtering step using TD-DFT calculations in Gaussian \cite{Jacquemin_Mennucci_Adamo_2011} was applied to estimate excitation wavelengths, oscillator strengths, and dipole moments. This step, together with an oscillator strength threshold ($>0.01$), served primarily to reduce redundancies within the selected set, yielding 19 structurally and photophysically distinct candidates for synthesis and experimental validation (Figure \ref{fgr:figure4}C). Importantly, the core photophysical diversity was already captured through the generative design and Tanimoto-based filtering steps, which effectively explored distinct fluorophore chemical spaces. The calculated excitation wavelengths showed reasonable correlation with experimental emission wavelengths from ChemFluor molecules ($R^2$ = 0.63; Supplementary Figure 3).

\subsection{Experimental Characterization of the Spectral and Structural Properties of AI-generated dyes}

\begin{figure}
  \includegraphics[width=\textwidth]{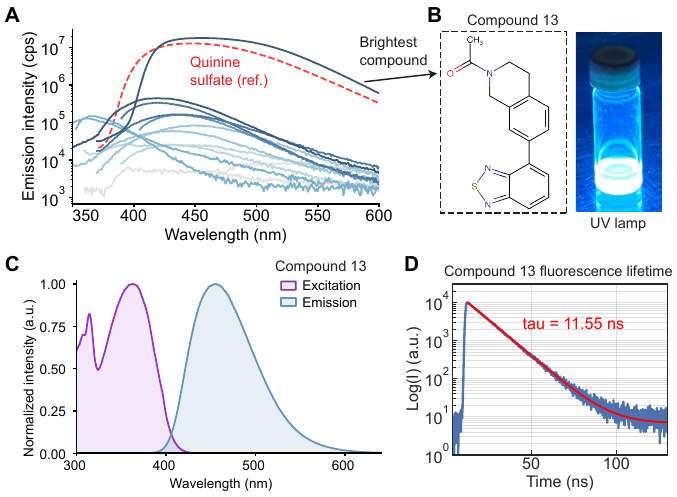}
  \centering
  \caption{\textbf{Experimental validation of fluorescent properties.} A) Emission spectra of 13 synthesized molecules compared to the quinine sulfate standard at 10 mM. B) Structure of the brightest compound (13) at 10 mM in chloroform under UV lamp excitation. C) Normalized excitation and emissions spectrum of compound 13 (Ex\textsubscript{max} = 363 nm, Em\textsubscript{max} = 460 nm). D) Fluorescence lifetime of compound 13 obtained from time-correlated single photon counting. The red line is a double exponential fit with mean lifetime of 11.55 s.
}
  \label{fgr:figure5}
\end{figure}

Fourteen of the 19 candidates were successfully synthesized by Enamine, though one was decomposed at time of receipt, leaving 13 molecules for experimental testing. Initial solubility tests indicated all 13 compounds were insoluble in DMSO and water at high concentration but were sufficiently soluble in chloroform for preparation of 10 mM solutions. Excitation and emission scans were performed to determine the spectra, Ex\textsubscript{max}, and Em\textsubscript{max} for the compounds, and these values were compared to quinine as a reference standard \cite{DROBNIK1966454} (Figure \ref{fgr:figure5}A, Supplementary Table 2).

Based on emission intensities, the three brightest compounds were identified as Compounds 13, 2, and 11, in descending order of brightness (Supplementary Figure 4). These compounds were predicted computationally to have high probabilities of  $PLQY > 0.5$ with values of 0.642, 0.830, and 0.861, respectively. Compound 13 had the highest emission intensity and was visibly the brightest upon UV excitation (Figure \ref{fgr:figure5}B); its excitation and emission spectra are shown in Figure \ref{fgr:figure5}C. Fluorescence lifetime measurements were performed on the top three brightest compounds using a 10 MHz pulsed nanoLED at 405 nm, with emission recorded at the respective Em\textsubscript{max} for each compound. Time-resolved fluorescence decay curves were tail-fitted to a bi-exponential model, yielding fluorescence lifetimes of 11.55 ns for Compound 13 (Figure \ref{fgr:figure5}D), 1.8 ns for Compound 2, and 1.5 ns for Compound 11 (Supplementary Figures 4-5). Remarkably, compound 13 displays a lifetime that is substantially longer than other blue-emitting dyes (DAPI (5 ns)\cite{dapi_lifetime}, Hoechst 33342 (3.4 ns)\cite{hoechst_lifetime}, Coumarin 343 (3 ns)\cite{coumarin_lifetime}, Atto 425 (3.6 ns)\cite{atto_alexa_lifetime}, Alexa Fluor 405 (1.5ns)\cite{atto_alexa_lifetime}, and Pacific Blue (2.1 ns)\cite{pacific_blue_lifetime}). Compound 13 also exhibited an unusually large Stokes shift of 97 nm, a feature that minimizes spectral overlap in imaging applications and makes it particularly well-suited for multiplexed imaging applications.  Further photophysical characterization of Compound 13 relative to quinine revealed a quantum yield of 0.62 and a molar extinction coefficient of 6000 M\textsuperscript{-1} cm\textsuperscript{-1}.

To evaluate the biological applicability of Compound 13, we conducted live-cell fluorescence microscopy at varying concentrations. HEK293 cells were incubated with Compound 13 at $0.1\,\mu\text{M}$, $1\,\mu\text{M}$, and $10\,\mu\text{M}$ then imaged in wide-field fluorescence (Figure 6). Cells exhibited minimal background in the no-dye control (Figure 6A), and progressively higher fluorescence with increasing concentration (Figure 6B-D). To quantify fluorescence intensity, mean pixel intensities were measured from fluorescence-only images (Supplementary Figure 6). As summarized in Table 1, the signal increased from 0.02 ($0.1\,\mu\text{M}$) to 0.121 ($1\,\mu\text{M}$) and at 0.140 ($10\,\mu\text{M}$). These results confirm that Compound 13 is cell-permeant and fluorescent under standard DAPI excitation, with strong signal-to-background contrast.  This dose-dependent profile, combined with its long fluorescence lifetime (11.55 ns) and large Stokes shift (97 nm), supports the utility of Compound 13 for high-resolution imaging applications requiring high photon turnover and compatibility with standard wide-field and confocal systems.

\begin{figure}
  \includegraphics[width=\textwidth]{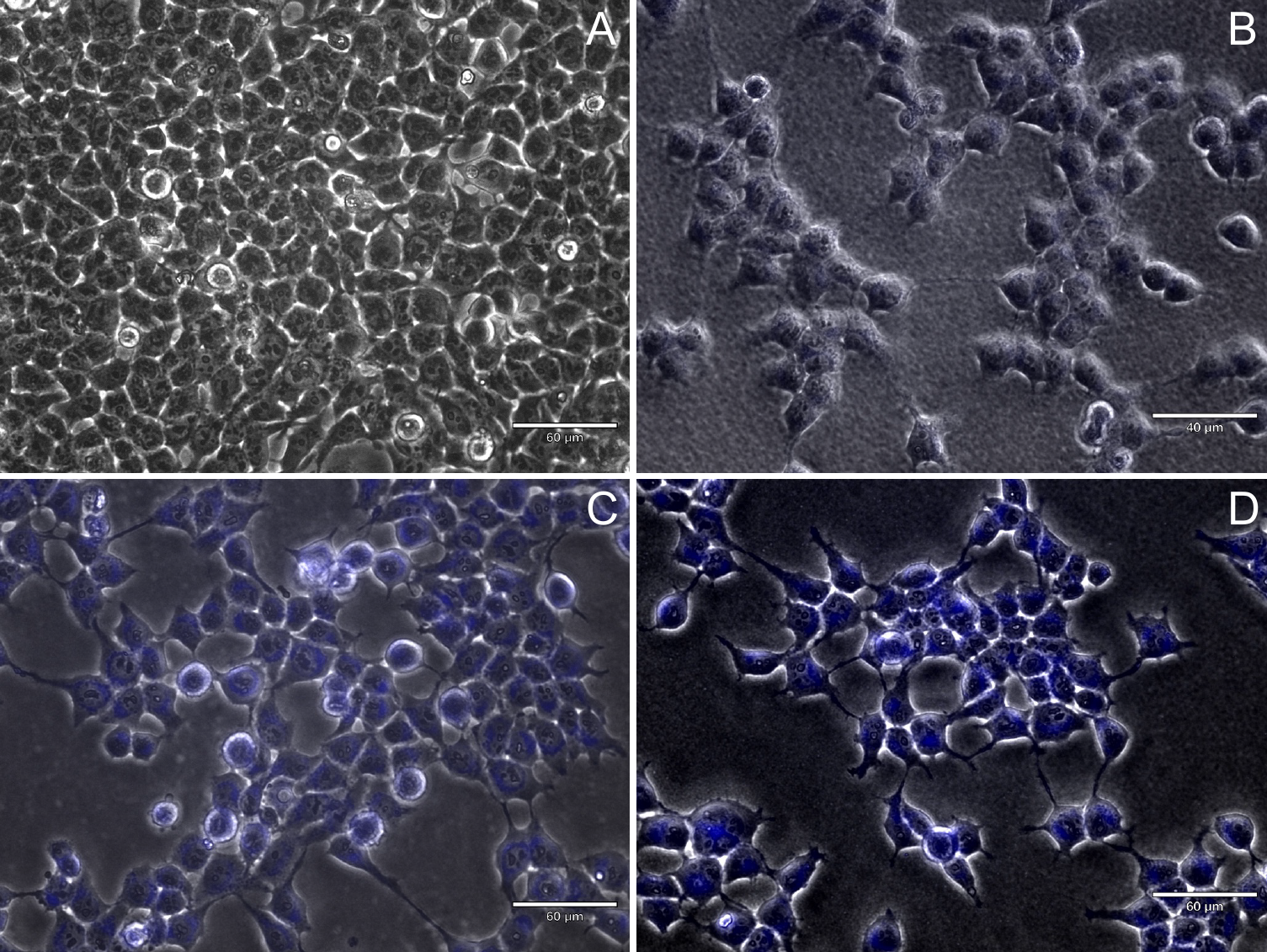}
  \label{fgr:figure6}
  \centering
  \caption{\textbf{Fluorescence of generated compound in cells.} Dose-dependent fluorescence of cells incubated with Compound~13. Representative images of cells treated with increasing concentrations of the dye candidate: (A) no dye (control), (B) 0.1\,$\mu$M dye, (C) 1\,$\mu$M dye, and (D) 10\,$\mu$M dye. Images are shown as overlays of the fluorescence signal (blue false color) on the corresponding transmitted light images. Scale bars, 60\,$\mu$m (A, C, D) and 40 \,$\mu$m (B).
}
\end{figure}

\section {Discussion}

In this study, we developed SyntheFluor-RL, a generative AI model capable of identifying diverse, readily-synthesizable fluorescent dye scaffolds from a vast chemical space of 32 billion molecules. We showed that three SyntheFluor-RL-generated molecules absorbed UV light and fluoresced in the visible range. These scaffolds had diverse chromophores, demonstrating the potential of SyntheFluor-RL for the design of new fluorescent molecules.

To accomplish our goals, we needed SyntheFluor-RL to robustly predict fluorescent properties. Because reinforcement learning value models require an RL value model architecture that is differentiable, we exclusively experimented with neural property predictor architectures, including graph neural networks and MLPs. Since PLQY, absorption, and emission properties depend on solvent, it was necessary to encode solvent features in our property predictor models. Doing so increased model expressivity and allowed us to train our model on more molecules, given that there exists no standardized experimental training data. Future improvement is likely to come from curating a larger fluorescence dataset, particularly one with a wider range of absorption and emission wavelengths.

Central to SyntheFluor-RL’s development from the core SyntheMol-RL algorithm is the adaptation of the multi-parameter objective from two antibiotic properties to four fluorescent properties: PLQY, absorption wavelength, emission wavelength, and sp$^2$ network size. Optimizing for these four parameters ensured that the candidate molecules possess the structural and electronic properties necessary for fluorescence. Additionally, to increase the number of promising candidate scaffolds generated by SyntheFluor-RL, we incorporated 57 new reactions, many of which have the potential to expand sp$^2$ networks. Eight of the fourteen molecules that were successfully synthesized used a reaction from this extended set, indicating the importance of including these reactions.

To demonstrate viability, we characterized 13 chemically stable compounds, identifying a core fluorophore scaffold that serves as a proof-of-concept for subsequent derivatization to enhance and extend established fluorescent properties. Of these 13, one compound (compound 13) was by far the brightest; however, 6 additional compounds showed fluorescence emission within two orders of magnitude of compound 13. The remaining six compounds were very weakly fluorescent. Notably, the chromophores of the three most strongly fluorescent molecules were all different. From brightest to dimmest, these molecules contained a benzothiadiazole, a benzofuran, and an isoxazolopyridine, respectively. These three molecules all fluoresce in a similar spectral region, but their lifetimes span an order of magnitude range. While benzofuran- and benzothiadiazole-based fluorophores have been previously described, these two derivatives have not previously been synthesized and evaluated for their photophysical properties \cite{belmonte2019synthetic,chen2023full, neto2022fluorescent,niu2015aminobenzofuran}. Both scaffolds have been seen to be highly tunable and photostable, which has resulted in their use in various bioimaging modalities. 

Compound 13 is an example of a blue-emitting fluorophore that integrates an unusually large Stokes shift (97 nm) with a long fluorescence lifetime (11.55 ns), surpassing those of six widely utilized UV-excited dyes, including DAPI, Hoechst 33342, Coumarin 343, Atto 425, Alexa Fluor 405, and Pacific Blue. Although its molar extinction coefficient (6,000 M$^{-1}$·cm$^{-1}$) is at least five-fold lower than these reference dyes, which typically exceed 30,000 M$^{-1}\cdot$cm$^{-1}$, this parameter is synthetically tunable through extended conjugation and scaffold modification. Compound 13 serves as a foundation for further development of blue-emitting dyes optimized for multiplexed imaging applications that benefit from large Stokes shifts and extended excited-state lifetimes. Notably, this compound also exhibits biologically relevant, live-cell-compatible properties, enabling high-contrast live-cell imaging without significant cytotoxicity. Importantly, owing to the modular design of SyntheFluor-RL, structurally related derivatives of these experimentally validated molecules can be readily generated to further tune and enhance their photophysical and biological performance.

While other AI-guided fluorescent molecule design approaches use property predictor models to filter promising candidates \cite{wang2021development,ye2020predicting,ju2021chemfluor, huang2024ai, bu2023designing}, these methods suffer from a limited library size and compounds that are not straightforward to synthesize. In contrast, our generative approach enables the design of compounds from broader chemical space \cite{tan2023novo}.

A related work, ChemTS, also used a generative fluorescent model (DNMG) \cite{sumita2022novo} and experimentally validated one novel fluorophore scaffold. This approach used random forest ML models and TD-DFT quantum chemistry calculations in the generative process. In contrast, SyntheFluor-RL uses neural ML models in its generative process, and TD-DFT calculations are only used as a final filtering step. SyntheFluor-RL generated 11,590 compounds in about 16.5 hours, using 32 CPUs and 1 GPU compared to 3,643 candidate molecules from DNMG using 1024 CPU cores over 5 days. By using TD-DFT calculations in the filtering process instead of within SyntheFluor-RL, our workflow reduces the computational cost and time required to generate candidate molecules. We believe that, as additional building blocks and reactions become available through the REAL Space framework, our faster generation strategy will allow for rapid screening of the increasingly vast chemical space.

Another notable generative approach to fluorescent molecule design is the FLAME framework, which experimentally validated sixteen candidates and found one novel compound with bright fluorescence \cite{zhu2025modular}. In their approach, they developed a graph neural network property predictor (FLSF) that utilized a 728-bit fingerprint specific for fluorescence and coupled it with the open-source generative framework REINVENT 4 that also uses reinforcement learning to generate candidates. However, a key strength of SyntheFluor-RL lies in its synthesis-aware approach in which scaffolds can be accessed from readily available building blocks and established reactions to enable facile experimental validation. While the FLAME framework generated 1 million candidates that were then filtered to a scaffold chosen due to its structural novelty and synthesizability, SyntheFluor-RL exclusively generates readily-synthesizable candidates. Furthermore, SyntheFluor-RL prioritized the design of both diverse and novel compounds, resulting in the top three candidates all containing distinct chromophores.

SyntheFluor-RL is the first generative AI model to design diverse, readily-synthesizable fluorescent scaffolds, enabling an easy path from AI-driven molecular design to experimental validation. Future work will enhance both its property predictors and its building block and reaction scoring capabilities for greater fluorescent property prediction and tunability.

\section{Methods}

\subsection{Dataset and processing}

We trained our models on the ChemFluor dataset \cite{ju2021chemfluor}, which contain 2,912 unique molecules dissolved in 63 different solvents, resulting in 4,336 unique molecule-solvent pairs. In addition to the SMILES strings that correspond to each molecule and the solvent the molecule is dissolved in, the dataset also contains experimentally derived PLQY values, absorption spectra, and emission spectra; and the solvent constants for polarizability (SP), dipolarity (SdP), acidity (SA), and basicity (SB), which are relevant in assessing the solvent effect \cite{catalan2009toward}.

Since absorption, emission, and PLQY measurements were not available for all entries, three separate training sets were curated for each prediction task. All molecule-solvent entries that contained the SMILES string, SP, SdP, SA, and SB entries, and the relevant measurement (either PLQY value, absorption wavelength, or emission wavelength) were included in the relevant dataset. For the 50 duplicate molecule-solvent pairs, the measurement of interest was averaged across identical entries. This resulted in 3,055 molecule-solvent pairs with PLQY measurements, 4,202 molecule-solvent pairs with absorption wavelengths, and 4,333 molecule-solvent pairs with emission wavelengths. 

\subsection{Property Prediction Architectures} 
We used four different property prediction architectures: (1) Chemprop-Morgan, (2) MLP-Morgan, (3) Chemprop-RDKit, and (4) MLP-RDKit. The Chemprop-Morgan model consists of the Chemprop graph neural network (GNN) model augmented with the Morgan fingerprint, which indexes the presence of specific substructures centered around each atom in the molecule. The model takes in a molecular graph representation of each training molecule as input, with atoms as nodes and bonds as edges. The GNN aggregates features such as atom and bond type for each atom and bond in the molecule through three message passing steps, creating vector representations of local neighborhoods in the molecule within the neural network layers. The 300-dimensional vector representation is concatenated with Morgan fingerprints, which are 2,048 bits and were calculated with a radius of 2 using the cheminiformatics package RDKit’s \texttt{GetMorganFingerprintAsBitVect} function. Four numerical solvent features (SP, SdP, SA, SB) corresponding to polarizability, dipolarity, acidity, and basicity are also concatenated to the feature vector. The combined feature vector is passed through a multilayer perceptron (MLP) with one hidden layer, with a final activation that is a sigmoid for classification tasks (PLQY) or a linear layer for regression tasks (absorption, emission). The Chemprop-RDKit model consists of the same GNN, but instead of the Morgan fingerprint, 200 molecular features computed by RDKit (and the 4 solvent features) are appended to the 300-dimensional vector output from the GNN and input to the MLP layer. The MLP-Morgan and MLP-RDKit models have the same architecture as the MLP layer in the corresponding Chemprop models but do not include the GNN; thus, each model takes either a 2,052-dimensional feature vector (Morgan fingerprint with solvent features) or 204-dimensional feature vector (RDKit and solvent features). All models were run with Chemprop version 1.6.1.

\subsection{sp$^2$ Network Size Algorithm} \label{subsec:sp2_section_ref}
\begin{algorithm}
\caption{sp$^2$ Network Size}
\label{sp2_section}
\begin{algorithmic}[1]
\State \textbf{Input:} molecule
\State \textbf{Output:} (integer) size of the largest connected component of sp$^2$ atoms in molecule
\State Compute $sp2\_atom\_idxs$, the indices of all atoms that are sp$^2$ in $mol$
\State Compute $sp2\_neighbors[atom\_idx]$, a dictionary that stores the neighbors of each atom that is sp$^2$ in $mol$
\State $visited\_global \gets \emptyset$ 
\State $max\_count \gets 0$ 

\Function{DFS}{$atom\_idx, visited\_local$}
    \State append $atom\_idx$ to $visited\_local$
    \State append $atom\_idx$ to $visited\_global$
    \For{each $neighbor\_idx$ in $sp2\_neighbors[atom\_idx]$}
        \If{$neighbor\_idx \notin visited\_local$}
            \State \Call{DFS}{$neighbor\_idx, visited\_local$}
        \EndIf
    \EndFor
    \State \Return {len(visited\_local)} 
\EndFunction

\For{each $atom\_idx$ in $sp2\_atoms\_idxs$}
    \If{$atom\_idx \notin visited\_global$}
        \State $max\_count \gets \max(max\_count, \text{DFS}(atom\_idx, \emptyset))$ 
    \EndIf
\EndFor

\State \textbf{Output:} $max\_count$
\end{algorithmic}
\end{algorithm}
The sp$^2$ network algorithm is a depth-first search (DFS) algorithm that finds the size of the largest connected component of sp$^2$ atoms in a molecule (Algorithm \ref{sp2_section}).

\subsection{SyntheMol-RL Architecture}
SyntheFluor-RL retains the core generative process of SyntheMol-RL, which uses a reinforcement learning (RL) value function (implemented as either a Chemprop or MLP model) to guide its selection of molecular building blocks to form a molecule. The value function learns to compute the expected property score of one or more building blocks, allowing SyntheFluor-RL to select combinations of building blocks that lead to promising full molecules.  SyntheFluor-RL optimizes for four properties simultaneously, which extends beyond the original application of SyntheMol-RL to antibiotic design, which only optimized for two properties. The reward function of SyntheFluor-RL evaluates the quality of full, generated molecules by scoring their photophysical properties, and these scores provide feedback to the RL value function to improve its ability to evaluate the quality of the molecule's component building blocks.

The SyntheMol-RL algorithm used by SyntheFluor-RL takes a chemical synthesis tree $T$ as input. Each node $N \in T$ has $N_{mol}$, which is a set of one or more molecular building blocks from the Enamine REAL Space. SyntheMol-RL defines a value function $V(N)$ on all nodes, which is a model that takes in the building blocks in a node's $N_{mol}$ as input and outputs a prediction of the property score. During each rollout, $V(N)$ is applied to all nodes created at a given step, and nodes are sampled proportional to $e^{V(N) / \tau}$, where $\tau$ is a temperature parameter that can be tuned to affect the RL policy's exploration or exploitation. After a molecule $m$ is constructed at the end of a rollout, it is scored by a weighted combination of $L$ property predictors, $M_{k}$ for $k \in \{1,...,L\}$ with weights $w_{k}$ for $k \in \{1,...,L\}$ to obtain the molecule's overall property score, $p(m) = \sum_{k=1}^{L} w_k * M_k (m)$.

For the RL value function $V(N)$, we employed MLP-Morgan models trained on the properties $PLQY >0.5$, absorption wavelength, emission wavelength, and sp$^2$ network size. $V(N)$ is a weighted combination of models where $V(N) = \sum_{k=1}^{L} w_k * Z_k (N_{mol})$, and $Z_1,...,Z_L$ are deep learning models and $w_1,...,w_L$ are the same property weights. After each rollout, the algorithm stores tuples of $(N, M_1(m), ..., M_L(m))$ for every node $N$ created along the path of nodes that ultimately led to molecule $m$, creating a training set of nodes and the property prediction scores of the final molecule created from them for $Z_1,...,Z_L$. The RL models are trained at set intervals over the rollouts to predict property prediction scores for a generated molecule based on the building blocks of a node in its path using a mean squared error loss, thus updating $V(N)$. For the reward function, we utilized three Chemprop-Morgan models trained to predict $PLQY > 0.5$, absorption wavelength, and emission wavelength, as well as the sp$^2$ network size algorithm depicted in Algorithm \ref{sp2_section}. These models were chosen because Morgan fingerprints consistently outperformed RDKit features in property prediction. The reward function provides a detailed assessment of each generated molecule's fluorescence-related properties.

Overall, this hybrid approach enables SyntheFluor-RL to rapidly explore combinations of building blocks using the MLP-Morgan value function followed by a more accurate evaluation of the generated molecules with the Chemprop-Morgan reward function.

\subsection{Dynamic Weighting}

SyntheFluor-RL utilizes the dynamic tuning mechanism of SyntheMol-RL to automatically adjust the RL temperature and property weights over time in order to optimize the generated molecules for both diversity and the four desired properties.

The RL temperature is important in defining the balance of exploration and exploitation and therefore the diversity of the generated molecules. The dynamic tuning method adjusts the RL temperature to obtain a molecule similarity of $\lambda^*$ on average during generation. We set the RL temperature target similarity $\lambda^* = 0.6$, which means that on average the Tanimoto similarity of a newly generated molecule to the most similar previously generated molecule is $0.6$.

Dynamic property weight tuning works by computing the average success rate on each rollout and adjusting the property weights based on the average rolling success rate. The success rate is determined by success thresholds for each property.  We defined the success thresholds as 0.5 for PLQY score ($p(PLQY > 0.5) \geq 0.5 $), binarized absorption score = 1 (where the binarized absorption score is 1 if the absorption wavelength predicted by the corresponding regression model is in the range of 420 to 750 nm, else 0), and binarized emission score (same criteria as the binarized absorption score). The success threshold for the size of the largest sp$^2$ network was $\geq12$.

\subsection{Filtering}

To narrow the generated molecules to the most promising candidates, we applied several rounds of filtering. First, using Algorithm \ref{sp2_section}, we filtered out molecules with a sp$^2$ network size smaller than 12. Next, we filtered molecules according to their predicted PLQY score, which is equivalent to $p(PLQY > 0.5)$ for a given molecule. We retained molecules where $p(PLQY > 0.5)\geq 0.5$. Then, we filtered the molecules by their predicted absorption and emission scores, respectively, as determined by the dynamic weighting algorithm. Specifically, each molecule was assigned a value of 1 if its predicted absorption was in the visible light range (420 to 750 nm), and 0 otherwise. The same criteria were used for emission. We retained molecules with both a binarized emission value of 1 and a binarized absorption value of 1.

We applied K-means clustering to group the remaining molecules into 100 clusters based on Morgan fingerprints with Tanimoto similarity to enable the selection of a diverse set of compounds. Finally, one of the authors manually reviewed each cluster to identify the most promising fluorescent structures. To ensure diversity among the selected molecules, at most one molecule was selected from each cluster. We requested a quote from Enamine for these 52 molecules, of which 34 were available for potential synthesis.

After the ML-based and similarity filtering and the Enamine screen reduced our candidates from 11,590 molecules to 34, we applied a minor final computational filter to help prioritize which compounds to synthesize. We adopted the TD-DFT methodology from Sumita et al.\cite{sumita2022novo}, using Gaussian software\cite{Jacquemin_Mennucci_Adamo_2011} to perform geometry optimization with the B3LYP functional and 3-21G* basis set, facilitating convergence by bypassing eigenvalue checks, calculating the full force constant matrix, and setting the maximum number of optimization cycles to 1000. The initial molecular coordinates were determined using an MMFF force field as calculated by RDKit, and the most stable conformations were selected. Solvent effects were simulated using the Self-Consistent Reaction Field (SCRF) approach, with water modeled as the implicit solvent to match the majority of molecules in the Chemfluor dataset. After geometry optimization, Time-Dependent Density Functional Theory (TD-DFT) calculations were performed to compute the electronic excited states, specifically the first five singlet states, providing estimates of excitation wavelengths, oscillator strengths, and dipole moments.

We reported the excited-state energy, oscillator strength, and dipole moment of the final optimization round, selecting molecules with oscillator strength above 0.01 for synthesis. This computational step removed 15 molecules from the 34 remaining candidates. These computational estimates served as a preliminary filter prior to experimental validation of the actual photophysical properties in various solvents.
\subsection{Synthesis and characterization}

Candidate fluorophores were synthesized by Enamine (Kyiv, Ukraine) and their identity and purity were confirmed by LC-MS. For photophysical characterization measurements, compounds were dissolved in chloroform to a concentration of 10 mM and diluted with chloroform as indicated.

\subsection{Measurements of excitation, emission, and fluorescence lifetimes}

Excitation and emission spectra were collected using a Fluorolog 3 spectrofluorometer (Horiba Jobin Yvon). The spectra of 10 mM of each molecule in chloroform were acquired through 1 nm increment wavelength scans with excitation and emission slit widths set to 4 nm and a 0.1 s integration time. For excitation, a tungsten lamp served as the source, while photon collection was obtained using a photomultiplier tube (PMT) and recorded as counts per second (CPS).

Fluorescence lifetimes were measured using a Horiba time-correlated single-photon counting (TCSPC) unit, equipped with a nanoLED405LH (pulse duration: 705 ps) operated at a repetition rate of 1 MHz. Emission signals were collected using a PMT across 4,096 channels with a total time span of 200 ns (0.055 ns/channel). The instrument response function (IRF) was recorded using a 1000-fold dilution of the Ludox-40 scattering solution obtained from Sigma, generating a FWHM of 660 ps. Fluorescence decay profiles were tail fit to a double exponential model in Python to determine the amplitude-weighted mean fluorescence lifetime ($\tau$).

\subsection{Quantum yield and molar extinction coefficient}

The fluorescent quantum yield ($\Phi_{f}$), defined as the ratio of photons emitted to photons absorbed by a fluorescent molecule, was determined using the relative method with quinine sulfate as the fluorescence standard ($\Phi_{std} = 0.62$). Emission spectra were collected for 20 $\mu$M quinine sulfate in 0.1 M sulfuric acid and 20 $\mu$M compound X in chloroform, using a 330 nm excitation wavelength. The absorbance of both solutions at 330 nm was approximately identical ($\sim$0.077). The relative quantum yield was calculated from the integrated emission spectra using the formula: $\Phi_{f} = \Phi_{std} \times \frac{I}{I_{std}} \times \frac{n^2}{n_{std}^2}$, where $I$ and $I_{std}$ are the integrated fluorescence intensities of the sample and standard, respectively, and $n$ and $n_{std}$ are the refractive indices of the solvents. Measurements were performed under identical conditions to ensure an accurate comparison with the reference standard.

The molar extinction coefficient ($\varepsilon$) was determined using the Beer-Lambert law: $A=\varepsilon cl$, where $A$ is the maximum absorbance, $c$ is the molar concentration of the solution and $l$ is the path length of the cuvette (1 cm). Absorbance was measured using a Beckman DU 640 spectrophotometer with a 1 cm quartz cuvette and compound titrations ranging from 20 to 500 $\mu$M. $\varepsilon$ was obtained from the slope of the concentration versus absorbance data using linear regression (y-intercept = 0).

\subsection{Cell Culture, Dye Labeling, and Imaging}
HEK293 cells were cultured in Dulbecco’s Modified Eagle Medium (DMEM) supplemented with 5\% fetal bovine serum (FBS) and maintained at 37°C with 5\% CO$_2$. Cells between passages 8–10 were seeded onto glass-bottom dishes (MatTek) and imaged at ~70\% confluency. For dye labeling experiments, culture media were aspirated and cells were gently washed once with phenol red–free, FBS-free imaging medium. Compound 13 was then added at the indicated concentrations (0.1 $\,\mu\text{M}$, 1 $\,\mu\text{M}$, or 10 $\,\mu\text{M}$), and cells were incubated at 37°C in a humidified incubator for 15 minutes. Live-cell imaging was performed immediately thereafter at room temperature. Microscopy was carried out using the Echo Revolve or Revolution microscope system with a 40× objective lens under standard DAPI illumination (excitation: 385/30 nm; emission: 450/50 nm). Overlay images were generated by combining transmitted light and fluorescence channels. A custom MATLAB script was used to quantify mean pixel intensity across fluorescence-only images.

\section*{Author Contributions}
R.S., K.C., and J.X. trained, executed, and analyzed all models, adapted the SyntheMol-RL codebase for fluorescence, drafted the computational portions of the manuscript, and made the associated figures. K.S. wrote supporting code and implemented the new chemical reactions. J.D collected experimental fluorescence data. S.C., J.Z., J.J., R.V.S., and K.S. supervised research. All authors edited the manuscript.

\section*{Data Availability}
Data used for this work including training data, the property prediction models, SyntheFluor-RL generation inputs, and the generated molecules are publicly available on Zenodo at \url{https://doi.org/10.5281/zenodo.18203970}.

\section*{Code Availability}
The code used in this work is available on GithHub at \url{https://github.com/swansonk14/SyntheMol}, including instructions for reproducing the results of this paper.

\section*{Acknowledgements}
J.J. acknowledges funding support from NIH (K99GM147609 and R00GM147609) and the Damon Runyon Cancer Research Foundation (DRG-2318-18). K.S. acknowledges the support of the Knight-Hennessy Scholarship and the Stanford Bio-X Fellowship.

\section{Competing Interests}
K.C. participated in internships at PostEra and Denali Therapeutics during the course of this work. J.X. worked at insitro during the course of this work.

\section{Tables}
\begin{table}[ht]
\centering
\caption{Dose-dependent fluorescence intensity of Compound 13.}
\label{tab:dose_response}
\renewcommand{\arraystretch}{1.2}
\begin{tabular}{lccc}
\hline
\multicolumn{1}{c}{\textbf{Dye}} & \textbf{0.1\,$\mu$M (B)} & \textbf{1\,$\mu$M (C)} & \textbf{10\,$\mu$M (D)} \\
\hline
Mean Pixel Intensity & 0.020 & 0.121 & 0.140 \\
\hline
\end{tabular}
\end{table}

\section*{Supplementary Information}
\renewcommand{\thefigure}{S\arabic{figure}}
\setcounter{figure}{0}
\renewcommand{\thetable}{S\arabic{table}}
\setcounter{table}{0}
\begin{figure}[H]
  \includegraphics[width=0.7\textwidth]{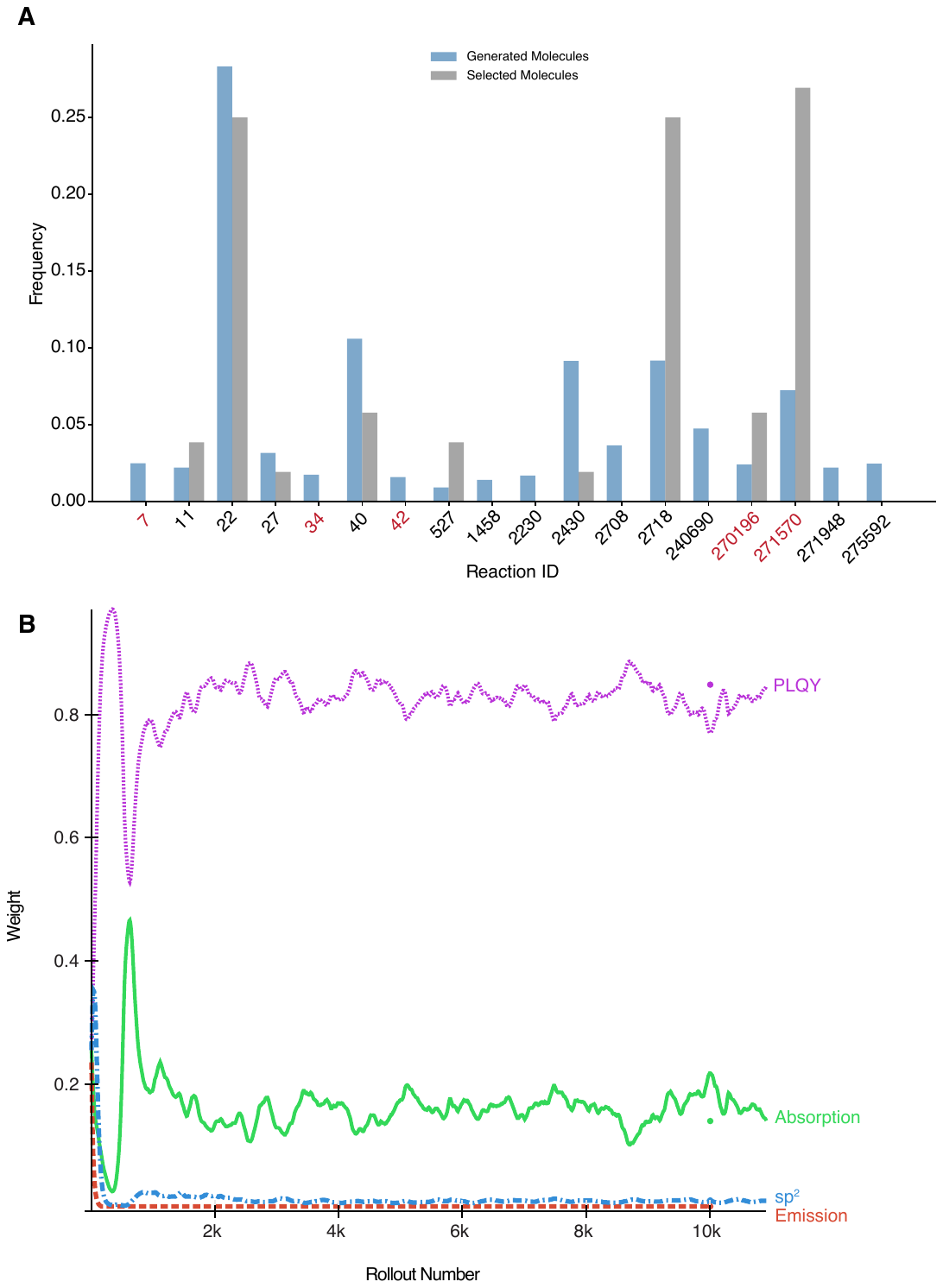}
  \caption{A) Histogram of reaction IDs used for generated molecules (blue) and selected molecules (gray). Reaction IDs in red are part of the extended set.
B) Line plot showing the change in weight associated with each MLP model (PLQY, absorption, emission, and sp$^2$) in SyntheFluor’s value function over the 10,000 rollouts.
}
  \label{fgr:supplementaryfigure1}
\end{figure}
\begin{figure}[H]
  \includegraphics[width=0.75\textwidth]{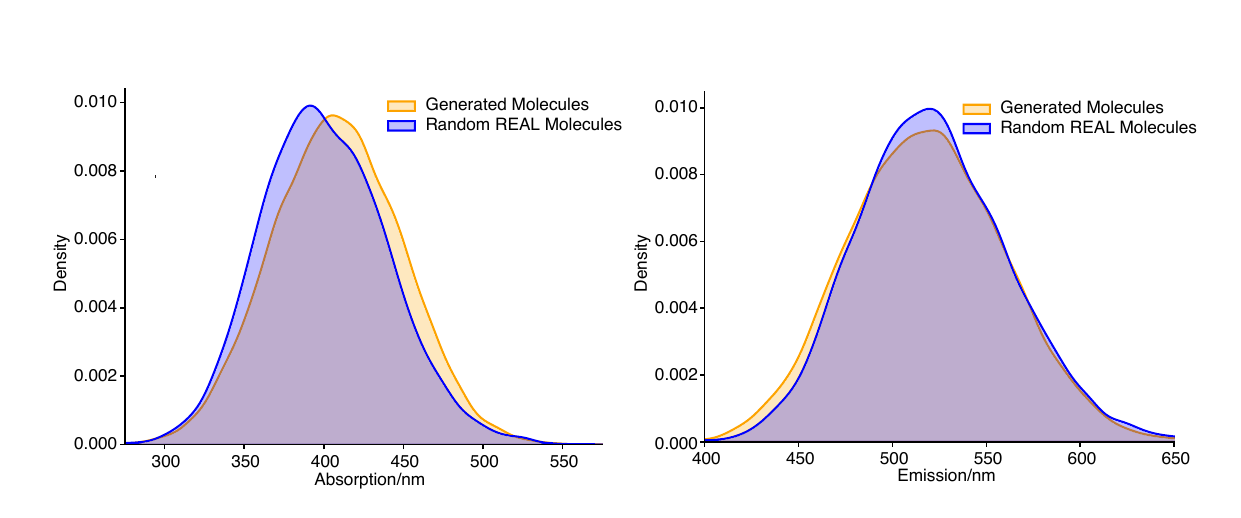}
  \caption{Distribution of Chemprop-Morgan predicted absorption wavelengths and emission wavelengths on generated molecules (yellow) versus a random sample of 10,000 molecules in the REAL Space (blue).
}
  \label{fgr:supplementaryfigure2}
\end{figure}
\begin{figure}[H]
  \includegraphics[width=0.75\textwidth]{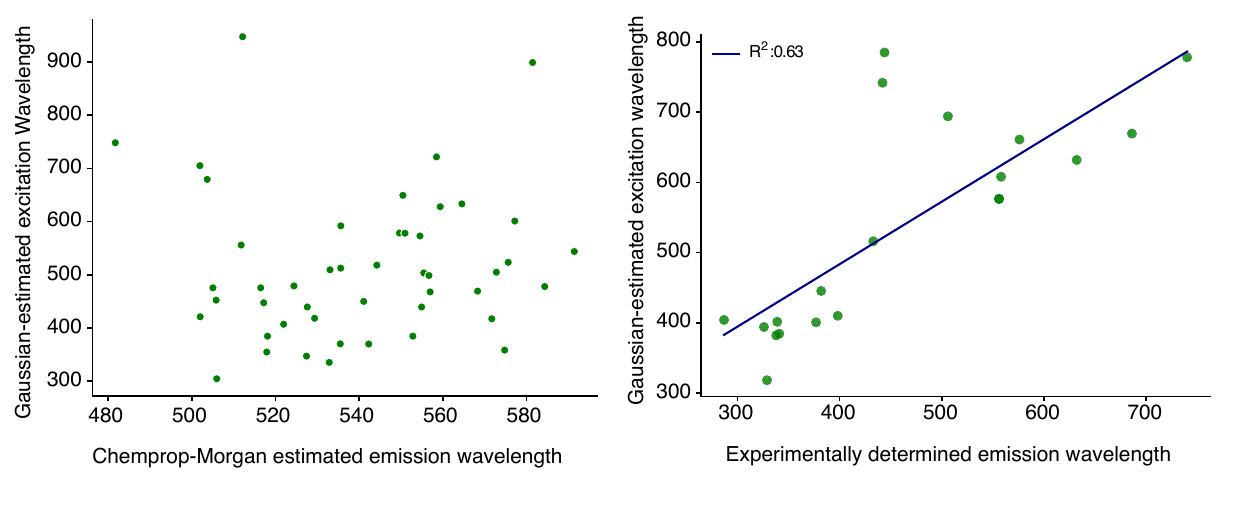}
  \caption{Plots of Gaussian-estimated excitation wavelength versus Chemprop-Morgan estimated emission wavelength among a subset of generated molecules (left), and of Gaussian-estimated excitation wavelength versus experimentally-determined emission wavelength among a subset of ChemFluor molecules (right).
}
  \label{fgr:supplementaryfigure3}
\end{figure}

\begin{figure}[H]
  \includegraphics[width=0.75\textwidth]{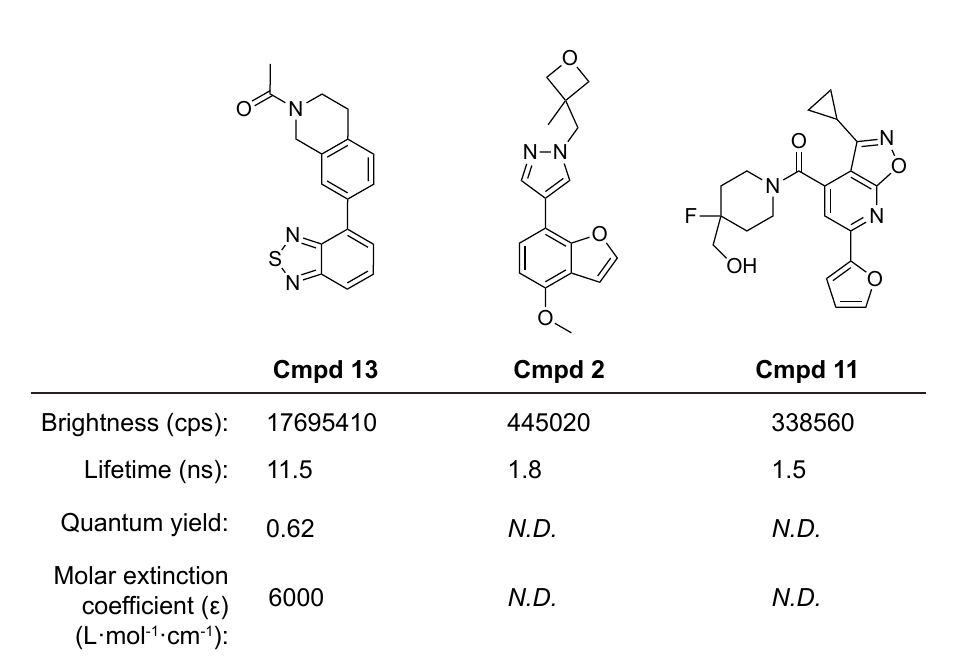}
  \caption{Fluorescent properties of the three brightest compounds generated by SyntheFluor.
}
  \label{fgr:supplementaryfigure4}
\end{figure}

\begin{figure}[H]
  \includegraphics[width=0.75\textwidth]{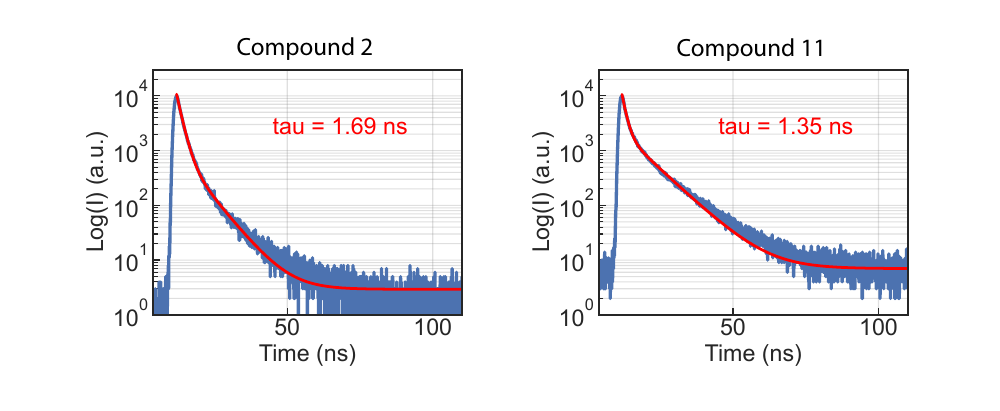}
  \caption{Fluorescence lifetime decay profiles of SyntheFluor compounds. Decay profiles are fit to a double exponential decay function (red curve) with amplitude-weighted mean lifetime tau.}
  \label{fgr:supplementaryfigure5}
\end{figure}

\begin{figure}[H]
  \includegraphics[width=0.75\textwidth]{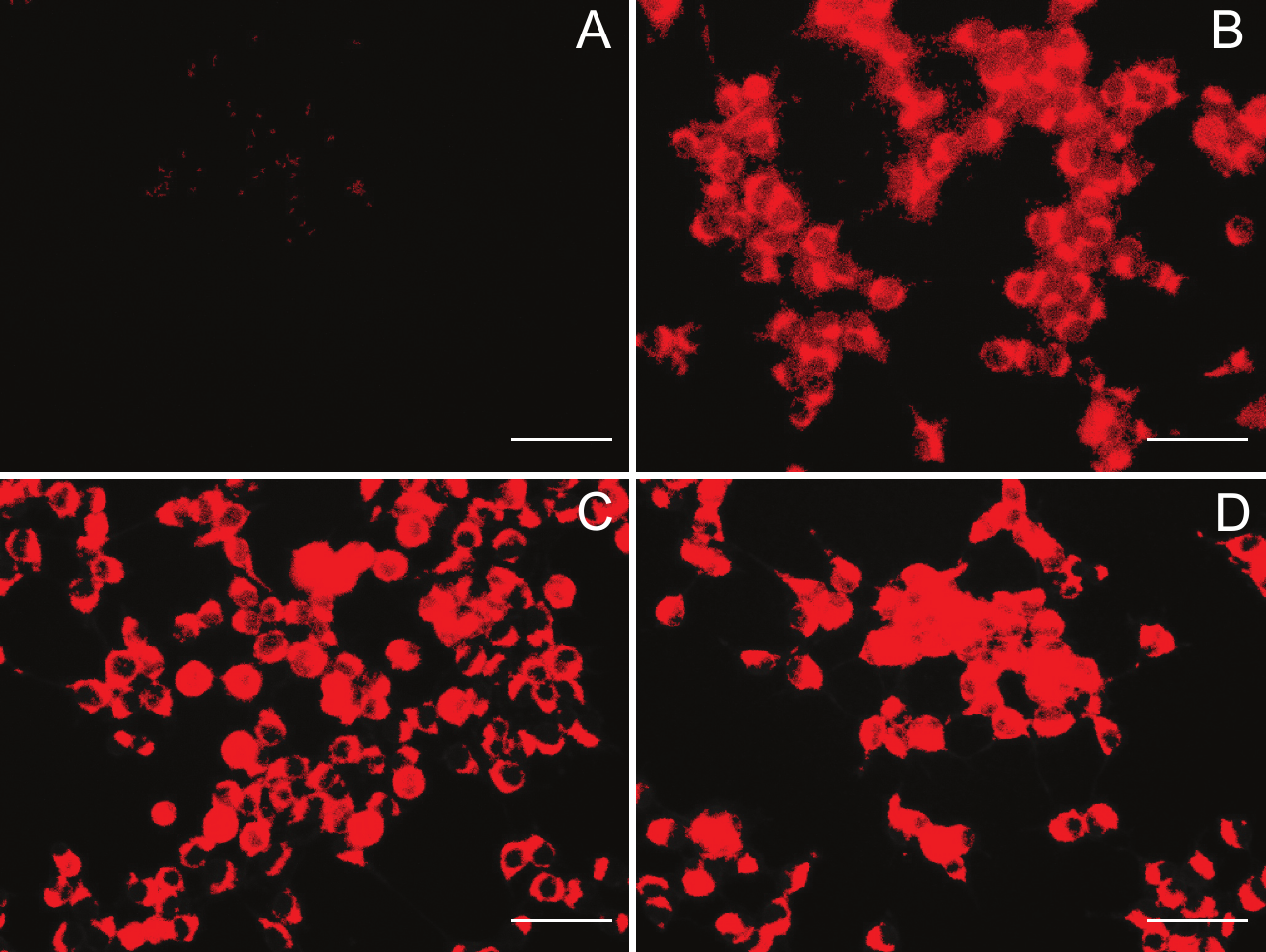}
  \caption{Fluorescence-only images used for quantification of mean pixel intensity.}
  \label{fgr:supplementaryfigure6}
\end{figure}

The remaining supplementary materials listed below are available on Zenodo at
\url{https://doi.org/10.5281/zenodo.18203970} organized under supplementary figure, table, and software folders.

Figure S6: UV/HPLC and mass spectrometry analysis of compound 13 

Figure S7: UV/HPLC and mass spectrometry analysis of compound 2

Figure S8: UV/HPLC and mass spectrometry analysis of compound 11

Table S1: Metrics for model performance on PLQY classification (Excel spreadsheet)

Table S2: 13 synthesized compounds ordered by brightness (CPS) (Excel spreadsheet)

Software S1: MATLAB script for fluorescence image quantification.

\end{document}